\documentclass{article}

\usepackage{arxiv}
\usepackage[numbers]{natbib}
\usepackage[utf8]{inputenc} 
\usepackage[T1]{fontenc}    
\usepackage{hyperref}       
\usepackage{url}            
\usepackage{booktabs}       
\usepackage{amsfonts}       
\usepackage{nicefrac}       
\usepackage{microtype}      
\usepackage{lipsum}

\usepackage[labelfont=bf]{caption}
\usepackage{graphicx}
\usepackage{graphbox}
\usepackage{graphics}
\usepackage{enumitem}
\usepackage{amssymb}
\usepackage{amsfonts,amssymb,amsthm}
\usepackage{float}
\usepackage{hyperref}
\usepackage{footnote}
\usepackage{mathrsfs}
\usepackage{amsfonts}
\usepackage{multirow}
\usepackage{bm}
\usepackage{makecell}
\usepackage{amsmath}
\usepackage{graphicx}
\usepackage{algorithm,algorithmicx,algpseudocode}
\usepackage{caption}

\usepackage{geometry}
\usepackage{amsthm}
\usepackage{makeidx}
\usepackage{csquotes}
\usepackage{booktabs}
\usepackage{adjustbox}
\usepackage{comment}
\usepackage{color}
\usepackage{amsfonts}
\usepackage{titlesec}
\usepackage[capitalise, noabbrev]{cleveref}
\usepackage{lineno}

\graphicspath{{figures2/}}

\newcommand{\R}{{\mathbb R}}
\newcommand{\N}{{\mathbb N}}

\newcommand{\E}{{\mathbb E}}
\newcommand{\cS}{{\mathcal S}}

\newcommand{\cQ}{{\mathcal Q}}

\newcommand{\w}{\ifmmode\bm{w}\else\textbf{\textit{w}}\fi}
\newcommand{\W}{\ifmmode\bm{W}\else\textbf{\textit{W}}\fi}
\newcommand{\Y}{\ifmmode\bm{Y}\else\textbf{\textit{Y}}\fi}
\newcommand{\X}{\ifmmode\bm{X}\else\textbf{\textit{X}}\fi}
\newcommand{\x}{\ifmmode\bm{x}\else\textbf{\textit{x}}\fi}
\newcommand{\y}{\ifmmode\bm{y}\else\textbf{\textit{y}}\fi}
\newcommand{\bb}{\ifmmode\bm{b}\else\textbf{\textit{b}}\fi}
\newcommand{\z}{\ifmmode\bm{z}\else\textbf{\textit{z}}\fi}

\newcommand{\ttt}{\ifmmode\bm{t}\else\textbf{\textit{t}}\fi}
\newcommand{\uu}{\ifmmode\bm{u}\else\textbf{\textit{u}}\fi}
\newcommand{\vv}{\ifmmode\bm{v}\else\textbf{\textit{v}}\fi}
\newcommand{\p}{\ifmmode\bm{p}\else\textbf{\textit{p}}\fi}
\newcommand{\D}{\ifmmode\bm{D}\else\textbf{\textit{D}}\fi}
\newcommand{\dd}{\ifmmode\bm{d}\else\textbf{\textit{d}}\fi}

\newcommand{\m}{\ifmmode\bm{m}\else\textbf{\textit{m}}\fi}
\newcommand{\M}{\ifmmode\bm{M}\else\textbf{\textit{M}}\fi}

\newcommand{\cc}{\ifmmode\bm{c}\else\textbf{\textit{c}}\fi}

\newcommand{\nn}{\ifmmode\bm{n}\else\textbf{\textit{n}}\fi}

\title{A Variational Auto-Encoder for Reservoir Monitoring}

\author{
  Kristian Gundersen \thanks{Corresponding Author: Kristian.Gundersen@uib.no} \\
  Department of Mathematics\\
  University of Bergen\\
  Bergen, Norway \\
   \And
  Seyyed A. Hosseini \\
  Bureau of Economic Geology\\
  University of Texas\\
  Austin, US \\
   \AND
   Anna Oleynik \\
   Department of Mathematics\\
   University of Bergen\\
   Bergen, Norway \\
   \And
   Guttorm Alendal \\
   Department of Mathematics\\
   University of Bergen\\
   Bergen, Norway \\
}

\begin{document}
\maketitle

\begin{abstract}
Carbon dioxide Capture and Storage (CCS) is an important strategy in mitigating anthropogenic CO$_2$ emissions. In order for CCS to be successful, large quantities of CO$_2$ must be stored and the storage site conformance must be monitored. Here we present a deep learning method to reconstruct pressure fields and classify the flux out of the storage formation based on the pressure data from Above Zone Monitoring Interval (AZMI) wells. The deep learning method is a version of a semi conditional variational auto-encoder tailored to solve two tasks: reconstruction of an incremental pressure field and leakage rate classification.  The method, predictions and associated uncertainty estimates are illustrated on the synthetic data from a high-fidelity heterogeneous 2D numerical reservoir model, which was used to simulate subsurface CO$_2$ movement and pressure changes in the AZMI due to a CO$_2$ leakage.
\end{abstract}


\section{Introduction}
Carbon Capture and Storage (CCS)  is a technology that captures CO$_2$, transport it and store it in suitable subsurface geological formations to isolate it from the atmosphere \citep{IPCC2005, ipcc2014mitigation}. To ensure that the captured CO$_2$ is retained within the targeted reservoir, proper monitoring of CCS sites is necessary \citep{Dixon:2015ee,Vermeul:2016cq}. Such monitoring is important for regulators, carbon trading \citep{Torvanger:2012bz}, and for public perception \citep{LOrangeSeigo:2011jp, Mabon:2014fj}. A challenge is to design monitoring programs \citep{JENKINS2015312,Alendal:2017gfa} based on model predictions \citep{Hvidevold:2015,Blackford:2020kf}. Information from the area we intend to monitor is usually only available at sparse location, e.g. monitoring wells. \\

In this study we will be focusing on Above Zone Monitoring Interval (AZMI) monitoring technique where one measures a property (in our case pressure) in a geological formation above the storage formation, with a limited number of strategically placed wells to look for changes as indicators of a potential migration of CO$_2$. This method has been previously used in many CCS field applications to monitor the possible CO$_2$ leakage \cite{hosseini2013static, kim2014above, ennis2017interpretation, coueslan2014integrated}. The core idea is to select an AZMI with minimal internal variations (changes are limited to natural cycles) where a physical or chemical property is in stable condition over time and ensure that there are no changes in that property as we start the injection operation in deeper formations. In this study we investigate if the pressure measurements in the AZMI can be used to not only detect, but also quantify a potential leakage. \\

We use a heterogeneous reservoir model to  simulate subsurface CO$_2$ movement scenarios, i.e. different migration fluxes and locations, and how it affects the pressure in the AZMI. Subsequently, these results are used in a deep learning framework that predict the pressure changes, measured only at the AMZI-wells, to the entire reservoir. \\

Instead of looking at the pressure itself, we found that incremental pressure, that is the pressure difference between two successive time steps, is better suited for our purpose. It stabilizes the time series by  eliminating mean and reducing influence from longer trends and potential seasonality. The pressure in the above zone is typically quite stable, and detection can be achieved by monitoring anomalies from the baseline. \\

Model snapshots, here individual incremental time steps of the entire domain from the simulations, are split into a training, validation and test data sets. A Convolutional Neural Network (CNN), with the ability to account for non-linear relationships, is trained to extrapolate from the monitoring wells to the entire field and to classify the associated flux rate of the leakage. The reconstruction enables us to estimate the spatial and temporal distribution of incremental pressure in the whole AZMI based on measurements in the wells, while the flux categorization gives an indication of the severity of the incident. \\

The reminder of the paper is outlined in the following matter. \cref{sec:ProblemFormulation} presents the formal problem to be solved. \cref{methods} presents the proposed method for solving the problem and necessary preliminaries. In the next section, \cref{Experimental_setup}, we shortly describe the simulations and how the data obtained from these simulations is preprocessed. In the same section we give the details related to the network model and optimization, and present the results. \cref{discussion} discusses the method and results, drawbacks, benefits and potential extensions. 

\section{Problem formulation} \label{sec:ProblemFormulation}
Let $p(s,t;\, par)$ denote the pressure in the AZMI. The pressure is obtained from computational fluid dynamics (CDF) simulations and depends on a location $s,$ time $t,$ and on a number of parameters, $par,$ such as, e.g., porosity, permeability, flux of leaked gas and boundary conditions. Assuming an uniform time grid we define an incremental pressure change as 
\begin{equation}\label{delta_p}
    \Delta p (s,t;\,par) = p(s,t+\Delta t;\, par) - p(s,t;\, par), \quad \Delta t>0.    
\end{equation}
Let simulation grid $\mathcal{S}=\{s_1,...,s_N\}$ consist of $N$ grid points $s_n,$ $n=1,...,N.$ Then $\Delta p (s,t;\,par)$ evaluated on $\mathcal{S}$ at a specific time $t$ and given parameter values can be arranged in a vector 
$\x^{(i)} \in \R^{N},$ i.e.,
\begin{equation}
\label{eq:x_i}
\x^{(i)}=(\Delta p(s_1,t; par),...,\Delta p(s_N,t; \, par))^T.
\end{equation}
Here the index $i$ corresponds to a specific time and parameter configuration. The collection of vectors $\x^{(i)}$ for all different times and parameter values, indexed with $i = 1,\dots K,$ constitutes the data set $\X.$ A location of unintended leakage and its flux are two of the parameters $par.$ Then a vector, or sometimes we refer to it as an instance,  $\x^{(i)}$ can be associated with a  leakage rate category $\gamma^{(i)} \in \{\gamma_1, \ldots, \gamma_r\},$ $r\in \N,$ with a certain probability $\mbox{Pr}(\gamma^{(i)} =\gamma_j),$ where $\sum_{j=1}^r\mbox{Pr}(\gamma^{(i)} =\gamma_{j})=1.$ Thus, we can introduce a category probability vector 
\begin{equation}
    \y^{(i)}=\left(\mbox{Pr}(\gamma^{(i)} =\gamma_{1}),\ldots,\mbox{Pr}(\gamma^{(i)} =\gamma_{r}) \right)^T, \quad \sum\limits_{j=1}^{r} (\y^{(i)})_j=1.
\end{equation}
If the leakage rate category is known, e.g., when using simulated data, $\y^{(i)}$ consists of zero entries except one which is equal to $1.$ The vectors $\y^{(i)} \in [0,1]^{r},$ $i=1,\ldots,K,$ constitute the set $\Y.$ The data set of pairs $(\x^{(i)}, \y^{(i)})$ is referred to as $\bm D.$ \\

\begin{figure}[h]
	\centering
		\includegraphics[width=0.60\linewidth]{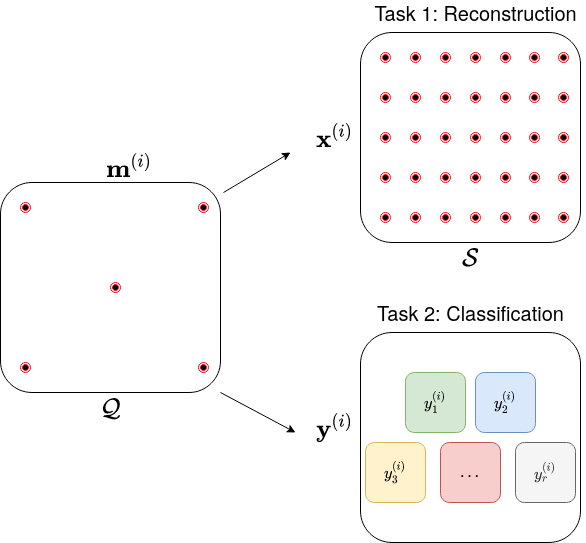}
		\captionof{figure}{Sketch of reconstruction of $\x^{(i)}$ and classification of $\y^{(i)}$ from $\m^{(i)}$. The upper right figure represent the grid $\cS$, and the left figure the subgrid $\cQ.$ The lower right figure represent the possible classes}\label{Sketch_map}
\end{figure}

The AZMI wells are assumed to be located at specific points in $\cS,$ that is, at $\cQ=\{q_1,...,q_M\} \subset \cS$ where $M$, the number of wells, is typically much less than $N.$  Hence, there is $\M=\{\m^{(i)} \in \R^{M}: \, \m^{(i)}=C\,\x^{(i)}, \, \forall \x^{(i)} \in \X \},$ where ${\bm C} \in \R^{M \times N}$ is a sampling matrix, 

\begin{align}\label{C_equation}
    ({\bm C})_{ij}= \left\{
    \begin{array}{ll}
        1,& \mbox{if } \, q_i =s_j\\
        0,& \mbox{otherwise}
    \end{array} \right. , \quad i=1,...,N \quad j=1,...,M.
\end{align}
The problem of reconstructing both the incremental pressure $\x^{(i)} \in \X$ and determining the associated $\y^{(i)}\in Y,$ from $\m^{(i)}\in \M,$ is presented in a schematic plot in \cref{Sketch_map}.

In this paper, we address the problem from a probabilistic point of view. Let $\x: \cS \to \R^{N}$ and $\m: \cQ \to \R^{M}$ be two multivariate random variables associated with the incremental pressure on $\cS$ and on $\cQ$, respectively, and $\y: \{\gamma_1,...,\gamma_r\}\to [0,1]^r$  a random variable describing probability of categories. Then the data sets $\X,$ $\M$ and $\Y$ consist of the realizations of $\x,$ $\m$, and $\y,$ respectively. We intend to approximate the probability distribution $p(\x,\y|\m)$ based on these data sets. Assuming that $\x$ and $\y$ are conditionally independent given $\m,$ the problem transforms into finding the distributions $p(\x|\m)$ and  $p(\y|\m).$ This formulation allows to predict $\x^{(i)}$ and $\y^{(i)}$ given $\m^{(i)}$ simultaneously, and to estimate an associated uncertainty of the prediction. To approximate the distributions, we  use a modified version of the Semi Conditional Variational Autoencoder \cite{gundersen2020semi}. 
\section{Methods}\label{methods}
The SCVAE is version of Conditional Variational Autoencoder developed in \cite{gundersen2020semi} for probabilistic reconstruction of flow state from sparse observations. Here we extend the SCVAE by including a classification task. Before we give details of the model, we provide a short overview of probabilistic autoencoders.

\subsection{Preliminaries}
Auto-encoders consist of two artificial neural networks coupled and trained together. The two parts are called encoder and decoder. The encoder takes $\dd \in \D$ as input and compress or encodes it into a latent representation $\z$. The decoder takes the latent representation as input and recreates $\dd$ given $\z$. \\

Auto-encoders have been used in a wide range of applications from dimension reduction \cite{hinton2006reducing}, anomaly detection \cite{sakurada2014anomaly, zhou2017anomaly} to drug discovery \cite{zhavoronkov2019deep} and machine translation \cite{cho2014properties}. These applications result in many different versions of auto-encoders, e.g., sparse  \cite{lee2008sparse, olshausen1997sparse}, contractive  \cite{rifai2011contractive}, denoising  \cite{vincent2008extracting} and variational auto-encoders \cite{kingma2013auto}. Auto-encoders can be divided into two groups: undercomplete and overcomplete, depending on the underlying architecture. An undercomplete auto-encoder has an architecture where the latent representation has a smaller dimension than the input. This structure forces the network to extract the most important features of the data and makes the auto-encoder suitable for dimension reduction. \\

Here we use a probabilistic version of undercomplete auto-encoder. Probabilistic auto-encoders are generalizations of traditional auto-encoders where the mappings $\dd \to \z$ and $\z \to \dd$ are stochastic. More specifically, let us assume that the data $\D$, is generated by a random process that involves an unobserved continuous random variable $\z.$ The process consists of two steps: (i) a value $\z^{(i)}$ is generated from a prior $p_{\theta^*}(\z);$ and (ii)  $\dd^{(i)}$ is generated from a conditional distribution  $p_{\theta^*}(\dd|\z^{(i)}).$  For convenience we assume that $p_{\theta^*}(\z)$ and $p_{\theta^*}(\dd|\z)$ come from parametric families of distributions $p_{\theta}(\z)$ and $p_{\theta}(\dd|\z),$ and their density functions are differentiable almost everywhere w.r.t. both $\z$ and $\theta$. Then $\dd^{(i)}$ can be reconstructed from a posterior distribution via \textit{latent representation} $\z \sim p_{\theta}(\z),$ that is,
\begin{equation} \label{eq:p_theta(x)}
\dd^{(i)} \sim p_{\theta}(\dd) = \int p_{\theta}(\dd|\z) d\z = \int p_{\theta}(\dd|\z)p_{\theta}(\z)d\z.
\end{equation}
In the right hand side of \cref{eq:p_theta(x)}, the latent representation $p(\z)$ is unknown, but can be approximated given the data, i.e. $p(\z|\dd)$. 
However,  $p(\z|\dd)$ cannot be easily obtained in neural networks with nonlinear activations, but can be approximated with a technique called variational inference \cite{blei2017variational}. The main concept of variational inference is to define a simple recognition model, say $q_\phi(\z|\dd)$, for $p(\z|\dd)$, which is paramatrized with the variational parameters $\phi$. Minimization of the Kullback-Leibler divergence \cite{kullback1959statistics} between $q_\phi(\z|\dd)$ and $p(\z|\dd)$
\begin{equation}
    D_{KL}[q_{\phi}(\z| \dd^{(i)})||p_{\theta}(\z| \dd^{(i)})] = \int q_{\phi}(\z|\dd^{(i)}) \log\left(\frac{q_{\phi}(\z|\dd^{(i)})}{p_{\theta}(\z|\dd^{(i)})}\right) d\z,
\end{equation}
with respect to $\phi$ gives an recognition model that approximates the true distribution \cite{kullback1951information}. The auto-encoder with the recognition model parameterized with variational parameters is called a Variational Auto-Encoder (VAE) \cite{kingma2013auto}. VAEs are typically used for generative modeling and allow for uncertainty estimation \cite{gundersen2020semi}.  \\

Here we use yet another version of VAE, the conditional variational encoder (CVAE), see e.g.  \cite{sohn2015learning}. The only difference from VAEs is that CVAEs allows for a conditional stochastic mappings from the data to the latent representation and opposite. That is the encoder and decoder are conditioned on a additional property, e.g. a label or a known function of the data. If the property is a known function of the data, as e.g. \cref{C_equation}, the CVAE simplifies to the semi-conditional variational auto-encoder (SCVAE) which was introduced in \cite{gundersen2020semi}. \\

As before we assume that the data $\D$ is generated by a random process that involves an unobserved continuous random variable $\z.$ In addition, we are given a random variable $\m,$ which can be observed, and whose observations form the data $\M$, see \cref{C_equation}. Then the prior and joined probability for $\dd$ are conditioned on $\m^{(i)},$ i.e., we have  $p_{\theta^*}(\z|\m^{(i)})$ and $p_{\theta^*}(\dd|\z^{(i)},\m^{(i)}).$ It is assumed that $p_{\theta^*}(\z|\m)$ and $p_{\theta^*}(\dd|\z,\m)$ come from parametric families of distributions $p_{\theta}(\z|\m)$ and $p_{\theta}(\x,\y|\z,\m),$ and their density functions are differentiable almost everywhere w.r.t. both $\z$ and $\theta$. \\

The neural networks are optimized in order to estimate $\theta$ and $\phi.$ A typical objective for the optimization is to maximize a log-likelihood function, $\log p_\theta(\dd^{(i)}| \m^{(i)}).$ When $d^{(i)}=(\x^{(i)},\y^{(i)})$ the aim could be to optimize $\log p_\theta(\x^{(i)}| \m^{(i)})$ and $\log p_\theta(\y^{(i)}| \m^{(i)}).$ To optimize more than one objective/task simultaneously is refereed to as  multi-task-learning (MTL) \cite{caruana1997multitask}. Below we show that these two aims coincide under conditional independence assumption on $\x$ and $\y.$ We give details of the SCVAE in the context of MTL and present the objective function below.

\subsection{Multi-task-learning in CCS monitoring}\label{MTL_SCVAE}
\begin{figure}[h]
    \centering
    \includegraphics[width=0.60\linewidth]{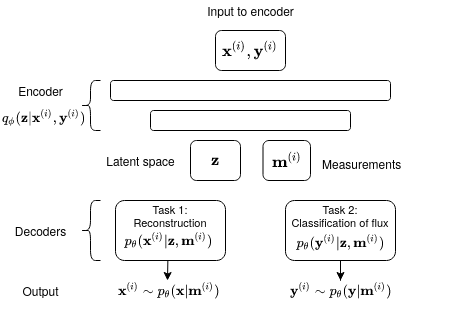} 
    \captionof{figure}{A sketch of the framework for the MTL-CCS-Monitoring algorithm.}\label{Model_v1}
\end{figure}

The SCVAE framework can be extended to include more than one tasks. Besides solving two tasks simultaneously the MLT-framework is beneficial as it reduces the the risk of overfitting \cite{baxter1997bayesian} and improves convergence properties \cite{caruana1997multitask}. Let the data $\D$ be now given by the pairs from $\X$ and $\Y,$ that is, $\dd=(\x, \y)$. Here the MTL-model consists of three parts; an encoder and two decoders, see a sketch of the model architecture in \cref{Model_v1}.

The latent representation $\z$ is approximated with a recognition model $q_\phi(\z|\x,\y)$ in the encoder. The decoders reconstruct and classify fluxes probabilistically, that is, approximate $p_\theta(\x|\m,\z)$ and $p_\theta(\y|\m,\z),$ respectively.  \cref{Model_v1} represents a MTL-framework with so called hard parameter sharing \cite{caruana1997multitask}. A large part of the hidden layers of the neural network is shared between the tasks. Here, the encoder contains the shared layers, and the two decoders contain the task specific layers. \\ 

We motivate this model choice as follows. As the overall aim  of the model is to maximize the log-likelihood $\log p_\theta(\x,\y|\m),$ the objective function can be obtained as the Evidence Lower Bound (ELBO) for the log-likelihood. It can be shown that  
\begin{equation}
    \begin{split}
         & \log p_{\theta}(\x^{(i)},\y^{(i)}|\m^{(i)}) \geq  \mathcal{L}(\theta,\phi;\x^{(i)},\y^{(i)},\m^{(i)}) \\ & = 
        \E_{q_{\phi}(\z|\x^{(i)},\y^{(i)})} \left[ \log p_{\theta}(\x^{(i)},\y^{(i)}|\z, \m^{(i)}) \right] 
        - D_{KL}[q_\phi(\z|\x^{(i)},\y^{(i)})||p_{\theta}(\z|\m^{(i)})].
    \end{split}\label{ELBO_dual_SCVAE}
\end{equation}
where $\mathcal{L}$ is referred to as ELBO \cite{vae_intro}. Instead of maximizing the log-likelihood, we maximize the ELBO, which is a common practice due to intractability of finding the true log-likelihood \cite{vae_intro}. Assuming that that $\x^{(i)}$ and $\y^{(i)}$ are conditionally independent given $\m^{(i)}$ and $\z,$ i.e.,
\begin{equation}
    p_{\theta}(\x^{(i)},\y^{(i)}|\m^{(i)}, \z)  = p(\y^{(i)}|\z, \m^{(i)})p(\x^{(i)}|\z, \m^{(i)}),\label{Conditional_independence}
\end{equation}
$\mathcal{L}$ in \cref{ELBO_dual_SCVAE} can be written as
\begin{equation}
    \begin{split}
            \mathcal{L}(\theta,\phi;\x^{(i)},\y^{(i)},\m^{(i)})  = &
        \E_{q_{\phi}(\z|\x^{(i)},\y^{(i)})} \left[ \log p_{\theta}(\y^{(i)}|\z, \m^{(i)}) \right]  \\ + &  \E_{q_{\phi}(\z|\x^{(i)},\y^{(i)})} \left[ \log p_{\theta}(\x^{(i)}|\z, \m^{(i)}) \right]   \\  
        - &D_{KL}[q_\phi(\z|\x^{(i)},\y^{(i)})||p_{\theta}(\z|\m^{(i)}).\label{final_dual_SCVAE_ELBO}
    \end{split}
\end{equation}
The ELBO in \cref{final_dual_SCVAE_ELBO} justifies the MTL-approach and motivate the choice of the model, \cref{Model_v1}. In order to take care of difference in scaling, prioritize different tasks, and possibly mitigate posterior collapse \cite{razavi2019preventing}, the terms in  \cref{final_dual_SCVAE_ELBO} could be scaled, i.e., 
\begin{equation}
    \begin{split}
            \mathcal{L}(\theta,\phi;\x^{(i)},\y^{(i)},\m^{(i)})  = &
        \E_{q_{\phi}(\z|\x^{(i)},\y^{(i)})} \left[ \log p_{\theta}(\y^{(i)}|\z, \m^{(i)}) \right]  \\ + &  \alpha \ \E_{q_{\phi}(\z|\x^{(i)},\y^{(i)})} \left[ \log p_{\theta}(\x^{(i)}|\z, \m^{(i)}) \right]   \\ 
        - &\beta   \ D_{KL}[q_\phi(\z|\x^{(i)},\y^{(i)})||p_{\theta}(\z|\m^{(i)}),\label{final_dual_SCVAE_ELBO_constrained}
    \end{split}
\end{equation}
where $\alpha, \beta > 0.$
This objective function can be obtained from constrained optimization formulation and corresponds to a lower bound for the Lagrangian under the KKT-slackness condition \cite{kuhn2014nonlinear, karush1939minima}. \\

Assuming $p(\z|\m^{(i)}) = \mathcal{N}({\bf 0}, {\bf I})$ and $q_{\phi}(\z|\x^{(i)},\y^{(i)}) = \mathcal{N}({\bm \mu}^{(i)}, \mathrm{diag}( \bm{\sigma}^{(i)})^2)$ we can express the KL-divergence term analytically

\begin{align}
    \begin{split}
    D_{KL}[q_{\phi}(\z|\x^{(i)}, & \y^{(i)})||p(\z|\m^{(i)})] = \\ & \frac{1}{2} \sum\limits_{j}^J  \left( \left(\sigma^{(i)}_{j}\right)^2 +  \left(\mu^{(i)}_j\right)^2 -  1 -   \log\left(\left(\sigma^{(i)}_j\right)^2 \right) \right).
    \end{split}\label{KL_div_compact}
\end{align}
where $J$ is the dimensionality of $\z.$ The expected log-likelihoods need to be estimated by sampling
\begin{equation}\label{eq:E-estimate}
    \begin{array}{l}
         \E_{q_{\phi}(\z|\x^{(i)},\y^{(i)})} \left[ \log p_{\theta}(\x^{(i)}|\z, \m^{(i)}) \right] \approx 
        \frac{1}{L}\sum\limits_{l=1}^{L} \log  p_{\theta}(\x^{(i)}|\z^{(i,l)},\m^{(i)}),\\
        \E_{q_{\phi}(\z|\x^{(i)},\y^{(i)})} \left[ \log p_{\theta}(\y^{(i)}|\z, \m^{(i)}) \right] \approx 
        \frac{1}{L}\sum\limits_{l=1}^{L} \log  p_{\theta}(\y^{(i)}|\z^{(i,l)},\m^{(i)}),\\
        \mbox{where } \, \z^{(i,l)} = g_{\phi}(\bm{\epsilon}^{(i,l)}, \x^{(i)},\y^{(i)}), \quad 
        \bm{\epsilon}^{l} \sim p(\bm{\epsilon}).
    \end{array}
\end{equation}
Here $\bm{\epsilon}^{l}$ is an auxiliary (noise) variable with independent marginal $p(\bm{\epsilon})$, $\mathcal{L}$ is the number of samples and $g_{\phi}(\cdot)$ is a differentiable transformation of $\bm{\epsilon},$ parametrized by $\phi,$   for details see \cite{kingma2013auto}. We can thus optimize

\begin{align}
    \begin{split}
        & \widehat{\mathcal{L}}(\theta, \phi, \x^{(i)},\m^{(i)},\y^{(i)})  =   \frac{1}{L}\sum\limits_{l=1}^{L} \log  p_{\theta}(\x^{(i)}|\z^{(i,l)},\m^{(i)}) \\
         &+ \alpha \frac{1}{L}\sum\limits_{l=1}^{L} \log p_{\theta}(\y^{(i)}|\z^{(i,l)},\m^{(i)}) -\beta D_{KL}[q_{\phi}(\z|\x^{(i)},\y^{(i)})||p_{\theta}(\z|\m^{(i)})].
        \label{Loss_function}
    \end{split}
\end{align}
This objective function $\mathcal{L}$ can be maximized with gradient decent method, however, usually it is to computationally costly to calculate the gradient over the entire data set. Thus we calculate the gradients on mini-batches instead, i.e. we use a stochastic gradient decent approach \cite{kiefer1952stochastic, robbins1951stochastic},

\begin{align}
    \begin{split}
    \widehat{\mathcal{L}}(\theta, \phi; \X, \M, \bm{Y}) \approx & \ \widehat{\mathcal{L}}^{R}(\theta, \phi; \X^R, \M^R, \bm{Y}^R) \\  = & 
    \frac{K}{R}\sum\limits_{r=1}^{R} \widehat{\mathcal{L}}(\theta, \phi; \x^{(i_r)}, \m^{(i_r)}, \y^{(i_r)}), \quad \alpha \geq 0.\label{obj_function_SCVAE_MTL}
    \end{split}
\end{align}
Here  $\X^{R}=\left\{\x^{(i_r)}\right\}_{r=1}^{R},$  $R<K$ is a minibatch consisting of randomly sampled datapoints, $\M^{R}=\left\{\m^{(i_r)}\right\}_{r=1}^{R}$ and $\bm{Y}^{R}=\left\{\y^{(i_r)}\right\}_{r=1}^{R}.$ After the network is optimized, a posterior predictive distributions $p_\theta(\x|\m)$ and $p_\theta(\y|\m)$ can be approximated with a Monte Carlo estimator.
\subsection{Uncertainty quantification}
Let  $\hat{\theta}$ and $\hat{\phi}$ be an estimation of generative and variational parameters of the models. Then the decoders can be used to predict the posterior as
\begin{align}\label{post_x}
    p_{\hat{\theta}}(\x|\m^*) \approx \frac{1}{N_{MC}} \sum_{j=1}^{N_{MC}} p_{\hat{\theta}}(\x|\z^{(j)},\m^*) \xrightarrow[N_{MC} \rightarrow \infty]{} \int
    p_{\hat{\theta}}(\x|\z, \m^*)p_{\hat{\theta}}(\z|\m^*)d\z.
\end{align}
and
\begin{align}
    p_{\hat{\theta}}(\y|\m^*) \approx \frac{1}{N_{MC}} \sum_{j=1}^{N_{MC}} p_{\hat{\theta}}(\y|\z^{(j)},\m^*) \xrightarrow[N_{MC} \rightarrow \infty]{} \int
    p_{\hat{\theta}}(\y|\z, \m^*)p_{\hat{\theta}}(\z|\m^*)d\z.\label{post_y}
\end{align}
From \cref{post_x} we can approximate the mean of the posteriori predictive distribution $\hat{\x}^*$ and empirical covarience matrix $\widehat{{\bm \Sigma}}_{\x}$ using  a Monte Carlo estimator. We get
\begin{align} 
      \widehat{\x}^*=
      \frac{1}{N_{MC}} \sum\limits_{j=1}^{N_{MC}} \x^{(j)}\quad \mbox{and} \quad
      \widehat{{\bm \Sigma}}_{\x} =  \frac{1}{N_{MC}-1} \sum\limits_{j=1}^{N_{MC}}(\x^{(j)} -  \widehat{\x}^*)(\x^{(j)} -  \widehat{\x}^*)^T,
        \label{eq:mean_and_cov} 
\end{align}
where $\x^{(j)}  \sim p_{\hat{\theta}}(\x|\m^*).$ The empirical standard deviation can be obtained by $\widehat{\bm \sigma}_{\x} = \text{diag}(\widehat{{\bm \Sigma}}_{\x})^{1/2}.$ Similarly, from \cref{post_y} we can approximate the mean of the posterior predictive distribution of the flux rate  and the co-variance matrix   
\begin{align} 
      \widehat{\y}^*=
      \frac{1}{N_{MC}} \sum\limits_{j=1}^{N_{MC}} \y^{(j)}\quad \mbox{and} \quad
      \widehat{{\bm \Sigma}}_{\y} =  \frac{1}{N_{MC}-1} \sum\limits_{j=1}^{N_{MC}}(\y^{(j)} -  \widehat{\y}^*)(\y^{(j)} -  \widehat{\y}^*)^T,
        \label{y_mean_and_std} 
 \end{align}
where $\y^{(j)}  \sim p_{\hat{\theta}}(\y|\m^*).$ The standard deviation can be found by $\widehat{\bm \sigma}_{\y} = \text{diag}(\widehat{{\bm \Sigma}}_{\y})^{1/2}.$ Confidence regions can be estimated as in \cite{gundersen2020semi}. \\

The vector $\widehat{\y}^*$ is the outcome of a softmax-function \cite{goodfellow2016deep_softmax} and is a probability that $\widehat{\x}^{*}$ corresponds to the $r$ different categories. The category that have the highest probability, i.e., $\gamma_{j^*}$ where $j^*=\arg\max\limits_{j} \widehat{\y}^*_j$, will be typically prescribed to the prediction $\widehat{\x}^*$.  \\

We will in \cref{Results_classification} use this argumentative maximum later and present results in form of a confusion matrix \cite{stehman1997selecting} to present the classification capabilities of the method.  
\section{Experiment}\label{Experimental_setup}

\subsection{Simulations}
\begin{figure}[t]
    \centering
    \includegraphics[width=0.65\linewidth]{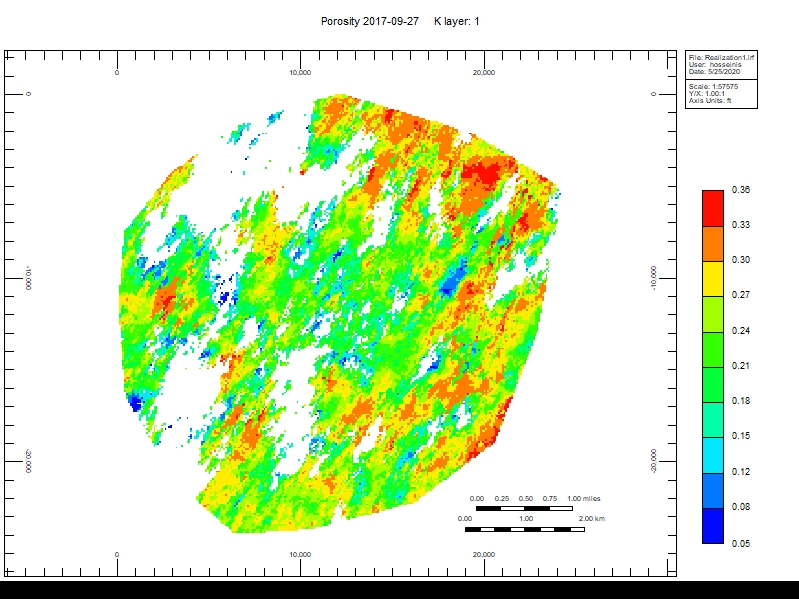}
    \includegraphics[width=0.65\linewidth]{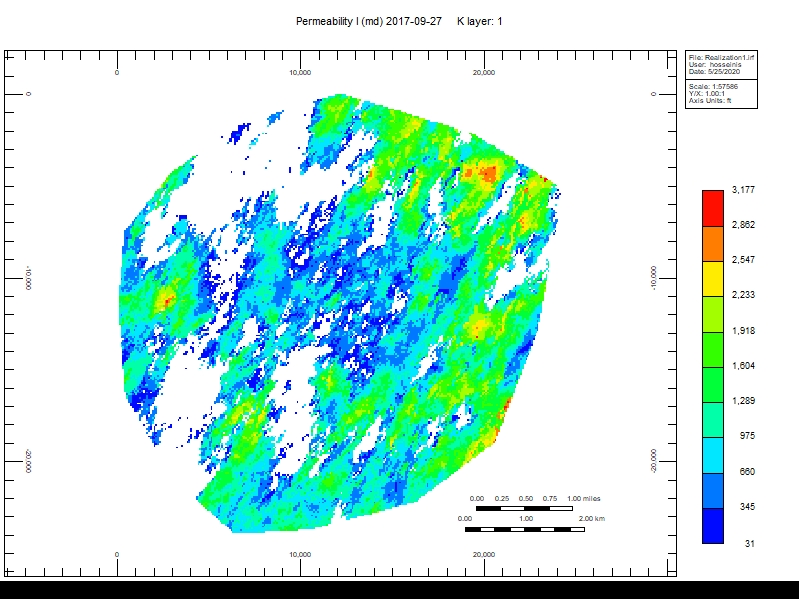}
    \caption{Screen dumps of the porosity and permeability used in the experiment retrieved from the CMG-software. \textbf{Upper Panel:} Porosity \textbf{Lower Panel:} Permeability}
    \label{fig:porosity_permability}
\end{figure}

The data for training and testing of the deep learning algorithm was generated by running 
dynamic numerical simulations on a static geologic model on a typical Frio Formation in Gulf of Mexico. The single layer model, with the layer's thickness $3$m, covers an area of approximate size of $7.5$ km by $7.5$km with grid resolution of $30$m x $30$m and has sand and shale facies where grid cells with shale facies are nullified. The simulation grid size is $486 \times 478.$ \cref{fig:porosity_permability} is showing the permeability and porosity distribution of this model. \\ 

This geological model is then used to run dynamic numerical simulation scenarios for fluid leakage using the Computer Modeling Group (CMG) software . The output of the simulations are pressure data in the AZMI. Leakage scenarios consist of five monitoring locations and one leaky well location. Location of the leaky well is varying in four different scenarios, while locations of monitoring wells are fixed. In addition, leakage rate is varied at four different levels as given in Table \ref{tab:Train_val_test_split}. The model outputs the daily mean of the pressure in the AZMI over a period of approximately $2.5$ years, i.e. $1001$ time steps. This means that a total of $16$ unique simulation each containing $1001$ pressure fields where used to create data for input to the deep learning model. Porosity and permeability field in all the simulations are the same.
The observations wells are located at grid points $\{(350,175),(290,290),(320,250),(260,150),(175,250)\},$ and the grid points for the different leakage release locations are $\{(170,138),(190,108),(102,90),(152,120)\}$. 
\subsection{Data and preprocessing}\label{sec:Pre-processing of Data}
The pressure obtained from the simulations has been pre-pocessed to form an input to the auto-encoder. Firstly, we have reduced the dimension of the original simulations. The layer grid was down-sampled from $486 \times 478$ to  $160 \times 160$ grid points as follows: we picked every third grid point in both directions, resulting in a $160 \times 162$ grid, and then we remove the last two columns. Hence, the pressure was calculated on a new grid $\mathcal{S}$ which contains in total $N=25600$ grid points. The main reason for this downsizing of the data set is GPU memory constraints and optimization speed.\\

The wells locations $\mathcal{Q}$, in reference to $\mathcal{S},$ can be indexed as
\begin{equation}
    \mathcal{Q} = \{(117,58),(97,97),(107,87),(87,50),(58,83)\},\label{monitoring_wells} 
\end{equation}
where the first number corresponds to the vertical and the second to the horizontal grid cell number in $\mathcal{S}.$ Similarly, the grid cell index of the leaky well is at one of the following grid cells 
\begin{equation}
    \{(103,70),(96,90),(125,102),(109,82)\}. \label{leaky_wells}
\end{equation}
Secondly, we calculated the incremental pressure as in \cref{eq:x_i}. that could be arranged in a vector, see \cref{delta_p}. However, for a practical implementation $\x^{(i)}$ was shaped as $160 \times 160$ in order to apply convolutional layers, see \cref{model}. Applying the difference transform stabilizes the time series around a mean and thus eliminates trend and reduces seasonality. This can help to reduce the complexity of the auto-encoder architecture and the computational time spent on optimization. For each of the $16$ simulations scenarios we have $1000$ incremental pressure fields. \\

In the third preprocessing step we removed the vectors $\x^{(i)}$ that contained extreme values. The extreme values typically occur in the simulation during the first few time steps after the gas migration starts. They are linked to transients causes when introducing CO$_2$ into the simulation domain. The instance  $\x^{(i)}$  was excluded if the absolute value of the incremental pressure change in the spatial domain is larger than a value of $5$ psi. This resulted in the data set $\X$ of $K=15241$ instances. \\

From running the different scenarios, the leakage rate related to each scenario is known. We thus create a one-hot representation $\y^{(i)}$ corresponding to $\x^{(i)},$ $i=1,...,K.$ Finally, we created $\M$ as described in \cref{sec:ProblemFormulation} with $\mathcal{Q}$ given in  \cref{monitoring_wells}. \\

After these pre-processing steps we split the data into a test, train and validation data sets. The test set contains $20 \%$ data randomly selected,  $20\%$ the remaining data was used for the validation and $80\%$ for the training. Hence, train, validation and test data sets contain $9753$, $2439$ and $3049$ instances, respectively. We have summarized some key statistics of pre-processed data in Table \ref{tab:Train_val_test_split}. The spatially down-sampled AZMI pressure data are made publicly available at \cite{Hosseini@Dataset}.
\begin{table}[h]
    \small 
    \centering
    \begin{tabular}{|c|c|c|c|c|c|c|}
        \hline
         & \multicolumn{2}{|c|}{Training Data} & \multicolumn{2}{|c|}{Validation Data} & \multicolumn{2}{|c|}{Test Data} \\ \cline{2-7}
        \makecell{Release rate \\ (MMSCFD)} & Quantity & Percent & Quantity & Percent & Quantity & Percent \\ \hline
        $\gamma_1$=100 000  & 2544  & 26.0 \% & 629  & 25.8 \% & 811 & 26.6 \% \\ \hline
        $\gamma_2$=200 000  & 2528  & 26.0 \% & 629  & 25.8 \% & 753 & 24.7 \% \\ \hline
        $\gamma_3$=300 000  & 2369  & 24.3 \% & 605  & 24.8 \% & 736 & 24.1 \% \\ \hline
        $\gamma_4$=400 000  & 2312  & 23.7 \% & 576  & 23.6 \% & 749 & 24.6 \% \\ \hline
        \textbf{Total} & \textbf{9753}  & \textbf{64} \% & \textbf{2439}  & \textbf{16} \% & \textbf{3049} & \textbf{20} \% \\ \hline
    \end{tabular}
    \caption{Overview of the train, validation and test data sets for the different CO$_2$ release rates.}
    \label{tab:Train_val_test_split}
\end{table}
\subsection{Model implementation }\label{model}
Here we give  details on the architecture of the encoder and the decoders. The encoder consist of several convolutional layers \cite{lecun1998gradient, goodfellow2016deep}, with ReLu \cite{hahnloser2000digital, nair2010rectified} activation functions. The decoder for the reconstruction mainly consists of transposed convolutional layers \cite{noh2015learning}. The decoder for classification is simpler, and consists of dense or perceptron layers \cite{minsky2017perceptrons}. In \ref{Appendix_A} we have given a schematic overview of the details of the different parts of the MTL-SCVAE model.    
\subsubsection*{Encoder - Shared layers}
The first layer is a dense layer with ReLu activations and filter size 25600 that takes $\y^{(i)}$ as input. Then this representation is reshaped to a dimension of $160 \times 160 \times 1$, i.e., the same dimension as $\x^{(i)}$. This allows to concatenate $\x^{(i)}$ and $\y^{(i)}$. Two CNN-layers with kernel size and strides two and ReLu activation functions then follows. The two CNN layers have 32 and 64 filters, respectively. This results in a representation of $(40 \times 40 \times 64).$ The representation is flattened and the consecutive layer is a dense layer with 16 filters and ReLu activation. This dense layer is input to two new layers with dimension $k$, i.e. the latent dimension (Here we use a dimension of $2$). These layers represent $\phi$ or the mean and standard deviation of the variational parameters. The layers are both input to a new layer that estimate the KL-divergence term, i.e. the latent representation $q_\phi(\z|\x^{(i)}, \y^{(i)})$. Linear activation are used for the two dense layers. 
\subsubsection*{Decoder - Reconstruction}
The decoder takes the measurements $\m^{(i)}$, i.e. the AZMI wells incremental pressure data as input in addition to the latent representation $\z^{(i)}$, that is the output of the encoder. The measurements and latent representation is concatenated. After this concatenation layer, a dense layer with 102400 filters that consecutively is reshaped to $(40 \times 40 \times 64)$ follows. Then follows two transposed CNN layers with stride and kernel size two and ReLu activation with filter size 64 and 32, respectively. These transposed CNN layers allows to expand the shape to $(160 \times 160 \times 32).$ The last layer is yet another transposed CNN layer with only one filter. This allows to obtain the same shape as $\x^{(i)}$, that is $(160 \times 160 \times 1).$ For this last layer linear activation functions is used. This layer is the output of reconstruction decoder.    
\subsubsection*{Decoder - Classification}
The classification of the flux has a less complex architecture. The decoder takes both the measurement $\m^{(i)}$ and the latent representation $\z^{(i)}$ as input. The two inputs are concatenated. Then follows three dense layers with ReLU activation functions and filter size 128, 64 and 32, respectively. The output layer is dense layer with filter size four and softmax activation function \cite{goodfellow2016deep_softmax}. The filters of the dense layer represent the probability for a specific class and summarize to one.    
\subsubsection*{Optimization}


As the experiments indicate good results and convergence, we use the objective function in \cref{final_dual_SCVAE_ELBO}. The scaled objective function, see \cref{final_dual_SCVAE_ELBO_constrained}, could improve the results. However, the choice of the regularization parameters $\alpha, beta$ is not straightforward, see e.g., \cite{heydari2019softadapt}, and we do not pursue this approach here. \\

\begin{figure}[H]
    \centering
    \includegraphics[width=0.70\linewidth]{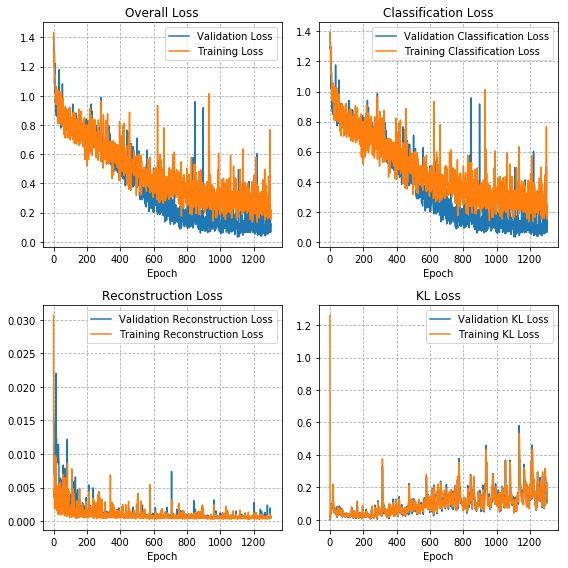}
    \caption{Convergence of the MTL-SCVAE optimization. \textbf{Upper Left:} Overall loss. \textbf{Upper right:} loss associated with the classification (here without scalling). \textbf{Lower Left:} The reconstruction loss. \textbf{Lower right:} The Kullback-Leibler loss.} 
    \label{convergence}
\end{figure}

For optimization we have used the ADAM algorithm \cite{kingma2014adam} which is a stochastic gradient decent method, similar to the RmsProp algorithm \cite{tieleman2012lecture}, but with adaptive momentum, i.e., the length and direction of the gradient decent step in the high dimensional objective function landscape. Further we have chosen a batch size of 128. To avoid over-fitting we used an early stopping regime, i.e., if we do not observe any improvement in the overall validation loss after $200$ epochs the optimization is stopped. In total the model run for $1305$ epochs, see \cref{convergence}. 
\subsection{Results}
This section is divided into two parts. One presents results related to reconstruction of the incremental pressure fields, and another one to classification of fluxes. We estimated the posterior predictive distribution as in \cref{post_x} and \cref{post_y}, with $N_{MC} = 100.$ The mean posterior and the standard deviation is calculated as in \cref{eq:mean_and_cov} and \cref{y_mean_and_std}.
\subsubsection*{Reconstruction}
\begin{figure}[t]
    \centering
    \includegraphics[width=0.65\linewidth]{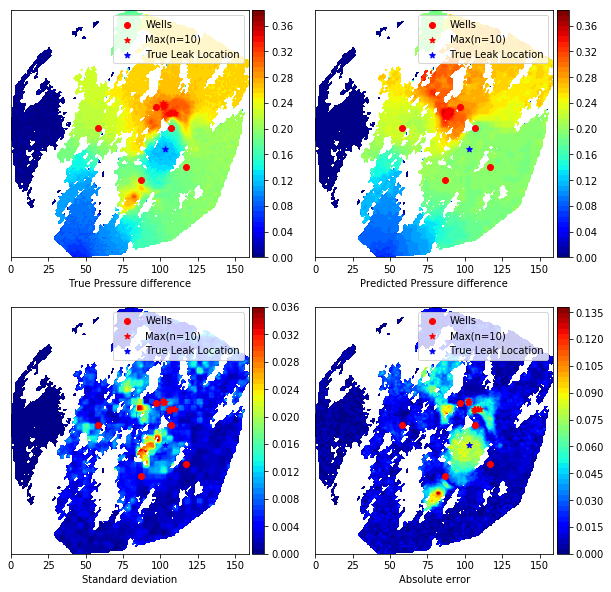}
    \captionof{figure}{Model dimension is 7.5 km by 7.5 km. \textbf{Upper Left Panel:} True incremental pressure \textbf{Upper Right Panel:} Reconstruction of the incremental pressure field. \textbf{Lower Left Panel:} Standard deviation of the reconstruction incremental pressure. \textbf{Lower Right Panel:} Absolute error between prediction and true incremental pressure field \textbf{Red circles:} AMZI-wells.  \textbf{Red stars:} Indices of the 10 largest incremental pressures. \textbf{Blue star:} True leak location. }\label{Reconstruction_results}
\end{figure}

\begin{figure}[t]
    \centering
    \includegraphics[width=0.70\linewidth]{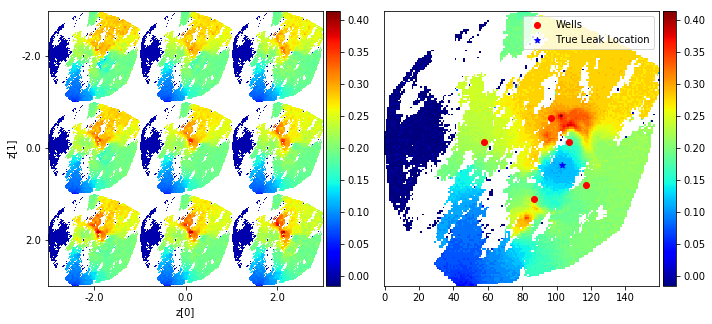}
    \captionof{figure}{The left panels shows different reconstructions  when sampling uniformly over $\z$. The right panel shows the associated true incremental pressure field.}\label{Sampling_different_versions}
\end{figure}

Figure \ref{Reconstruction_results} consists of $4$ different plots for different statistics of the instance number $i=422$ of the test data set. The first is the true incremental pressure, the second the mean of the posterior predictive distribution, the third the standard deviation of the predicted incremental pressure and the fourth the absolute error of the true and predicted incremental pressure field. We observe that the MTL-SCVAE model is able to reconstruct and predict the incremental pressure relatively well. The standard deviation is relatively high near where the incremental pressure change is the largest. The mean relative  $L_2$ error calculated as
\begin{equation}\label{L2_error}
  \mathcal{E} = \frac{1}{n}\sum\limits_{i=1}^{n}\frac{||\widehat{\x}^{(i)} - \x^{(i)}||_2}{||\x^{(i)}||_2}
\end{equation}
is equal to $\bf{0.1341}$.  Here $\x^{(i)},$ $i=1,\ldots,n$ correspond to the test data. This error indicates that the prediction error in average is about $13 \%$ of the true state. Further conclusions about leakage location, accumulation areas and uncertainties could be drawn from this prediction. 
Figure \ref{Sampling_different_versions} shows the true incremental pressure on the right and  $9$ different variations of the prediction, by sampling differently over $p_\theta(\z|\x^{(i)}, \y^{(i)})$. By varying $\z$ we can reconstruct statistical sound representations of the incremental pressure, given the measurements. We see that the prediction indeed varies the most in the regions with large standard deviation, see in \cref{Reconstruction_results} (Lower left). 

\subsubsection*{Classification}\label{Results_classification}
\begin{figure}[H]
    \centering
        \includegraphics[width=0.60\linewidth]{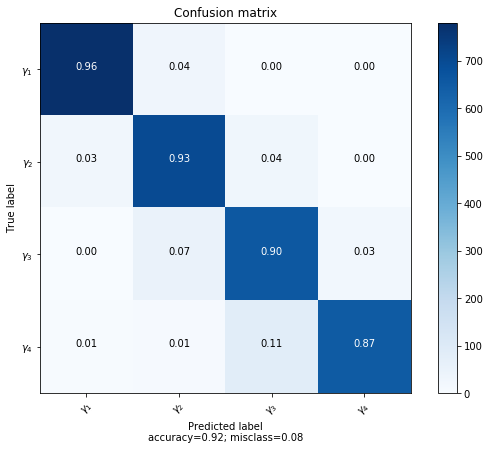}
    \captionof{figure}{Confusion matrix of the classification of fluxes}\label{confusion}
\end{figure}

Plotting the true positive rate versus the false positive rate at various thresholds we can create a Receiver Operating Curve (ROC) \cite{hajian2013receiver} and give insight about the models classification ability. 
We predict the posterior predictive distribution for each of the instances in the test data set such that we can create ROC that reflects the uncertainty in the classification.   
\begin{figure}[H]
    \centering
        \includegraphics[width=0.60\linewidth]{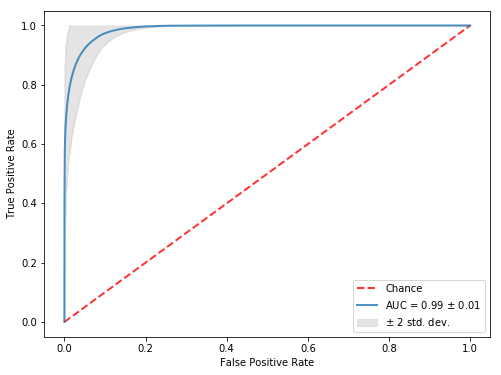} 
    \captionof{figure}{ROC of the classification of fluxes}\label{ROC_curve_figure}
\end{figure}
In Figure \ref{confusion} a confusion matrix of the multi-label classification of the different flux classes is presented. Here we use the argumentative maximum of $\widehat{\y}^{(i)}$ as predicted class. A confusion matrix gives a good overview of the models classification abilities and insight in which class-predictions the model has the greatest challenges with. From \cref{confusion}, the model miss-classifies most for leakage rates $\gamma_3$ and $\gamma_4,$ and mainly to nearby leakage rates. The overall classification accuracy is good. \\   

\begin{figure}[H]
    \centering
    \includegraphics[width=0.60\linewidth]{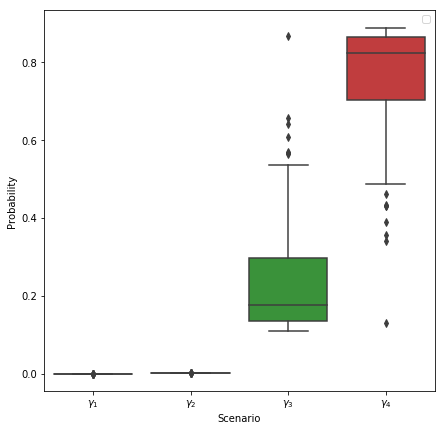}
    \captionof{figure}{Box plot for $100$ samples of the category's probability obtained for $422$ instance in the test data.}\label{fig:box_class}
\end{figure}

\begin{figure}[H]
    \centering
        \includegraphics[width=0.60\linewidth]{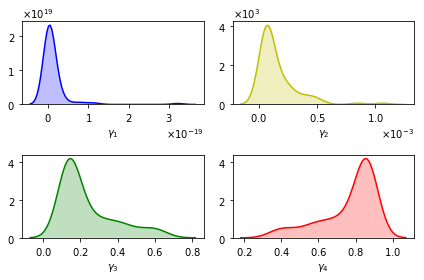}
    \captionof{figure}{The plot shows the predictive posterior distribution of the different leak categories $\gamma$ for sample 422. We have used the seaborn visualization library, i.e. Gaussian kernel with Scott method for estimation the kernel bandwidth. The KDE smoothens the empirical distribution, thus exceeding the estimate beyond the possible range of $[0,1]$.}\label{posterior_predictive_dist}
\end{figure}

\cref{posterior_predictive_dist} shows the posterior predictive distributions of the predicted leakage category/flux for sample number $422$ in the test data set. The figure shows that the model predicts leakage class $\gamma_4$ to be the most probable leakage class. This is in fact the true leakage category for this particular sample.   
\section{Discussion}\label{discussion}
We have presented the multi-task learning auto-encoder for reconstruction and classification based on sparse observations. The choice of the objective function for the network optimization is motivated from multi-task learning point of view and from classical ELBO derivation method. Conditioning the auto-encoder on observations and assuming conditional independence of two tasks allowed us to simplify the model structure, which is advantageous as it reduces the number of model parameters and requires less computational time.  \\

We have applied the method to reconstruct the incremental pressure field of the entire AZMI area and classified the flux of the leaks,  based on a limited number of measurements. In particular, we were able to distinguish between different leak categories with high confidence based on only measurements from $5$ wells.\\

The developed framework has the ability to quantify the uncertainties of the predictions for both  reconstruction and classification tasks. In case of a leak from a CCS reservoir, uncertainty estimation can be of crucial importance. Decision makers can make more informed decisions, implementing more effective and powerful countermeasures to mitigate the possible release of carbon dioxide to the atmosphere. \\

There are uncertainties related to input parameters to the simulator, especially with respect to permeability and porosity. Typically porosity and permeability are determined based on core samples at various locations in the reservoir and the well-known techniques are used to populate the entire domain. Here we have assumed a fixed porosity and permeability on the entire domain. A natural extension of our work would be to investigate an impact of uncertainity in geological heterogeneity. \\

A potential weakness of the presented experiment is that we have not  optimized hyper-parameters of the model systematically. Altering number of CNN-layers, the optimization procedure, kernel sizes, strides, activation function and regularization parameters might improve the results. We have not found it expedient to perform an extended hyper-parameter search, since the current configuration achieves good results.\\

Here we have performed the experiments on a 2D model. An extension to a 3D model is straight forward, but more computationally demanding and thus, not considered here. \\

An inherent difficult task is to find optimal locations of the AZMI-wells. The MT-SCVAE framework can be used to find locations that minimize the reconstruction error and flux rate classification. Here we used the model architecture where the measurements locations were fixed and not directly specified. Nonetheless, the auto-encoder was able connect the measurements to the locations which is indicated by better prediction fit closed to the observation wells. It is possible to modify the model architecture and link the measurements to their exact locations. While we have not seen much improvement in our experiments, this addition could be beneficial when using the model for optimizing the monitoring wells layout. 

\section*{Acknowledgements}
This work is part of the project ACTOM, funded through the ACT programme (Accelerating CCS Technologies, Horizon2020 Project No 294766). Financial contributions made from: The Research Council of Norway, (RCN), Norway, Netherlands Enterprise Agency (RVO), Netherlands, Department for Business, Energy \& Industrial Strategy (BEIS) together with extra funding from NERC and EPSRC research councils, United Kingdom, US-Department of Energy (US-DOE), USA. Kristian Gundersen has been supported by the Research Council of Norway, through the CLIMIT program (project 254711, BayMode) and the European Union Horizon 2020 research and innovation program under grant agreement 654462, STEMM-CCS. The authors would like to acknowledge NVIDIA Corporation for providing their GPUs in the academic GPU Grant Program. Authors also thank Gulf Coast Carbon Center at The University of Texas at Austin to provide the geological model for this study. Finally, we thank Nello Blaser for valuable comments and discussions during preparation of this manuscript. 
\clearpage
\appendix
\section{Details on the Experiment}\label{Appendix_A}
We use Keras \cite{chollet2015keras} in the implementation of the experiment. There is two extra dimension in the figures showing the encoders and decoders. That is, the implementation allows for more features and time steps. Here, these are one in each case, because we only consider one time step and one feature (incremental pressure).      
\subsection{Encoder}
\begin{figure}[h]
    \centering
    \includegraphics[width=0.85\linewidth]{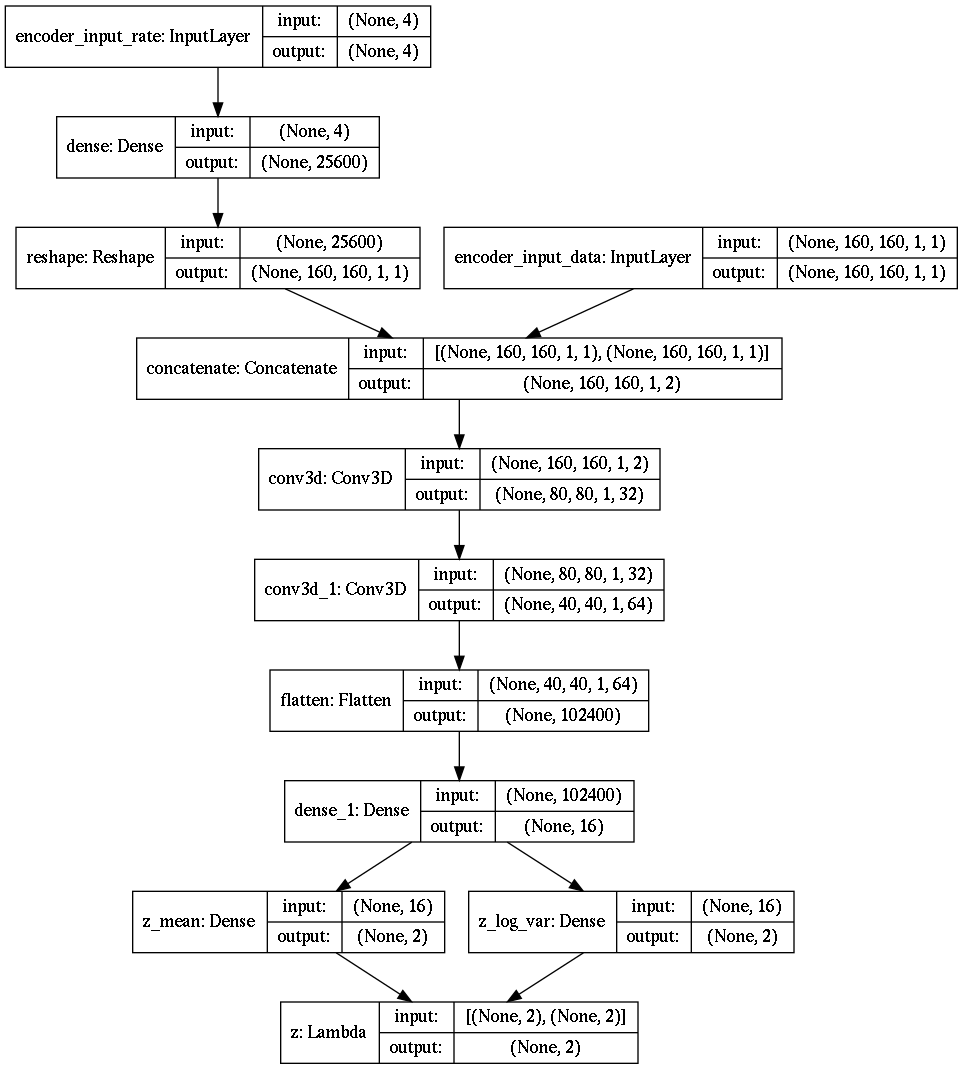}
    \caption{Architecture for the encoder}
\end{figure}
\clearpage
\subsection{Decoder - Reconstruction}
\begin{figure}[h]
    \centering
    \includegraphics[width=0.85\linewidth]{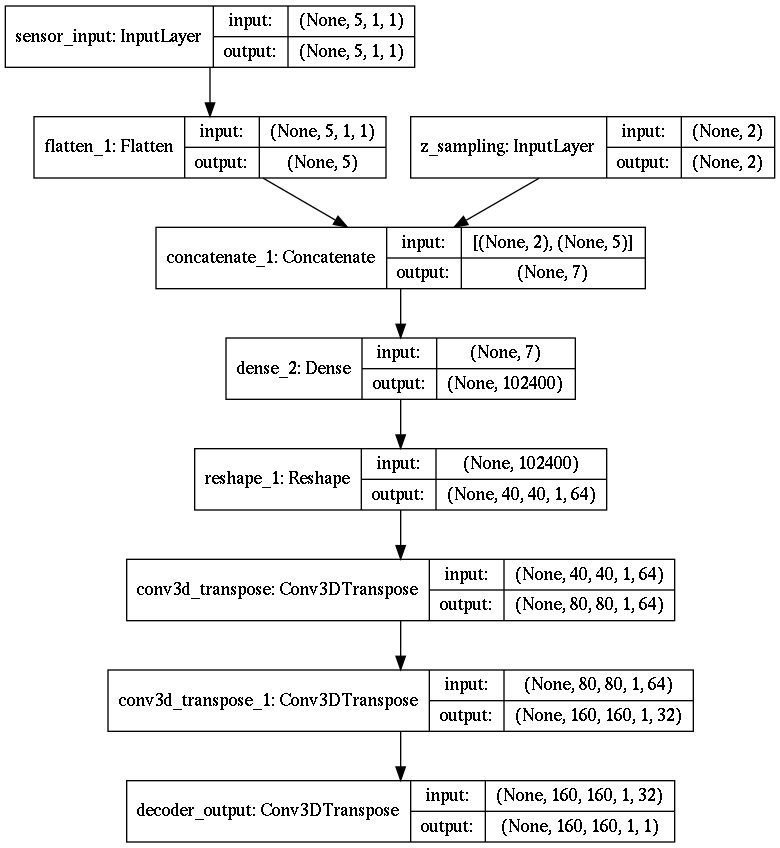}
    \caption{Architecture for decoder for the reconstruction task}
\end{figure}
\clearpage
\subsection{Decoder - Classification}
\begin{figure}[h]
    \centering
    \includegraphics[width=0.85\linewidth]{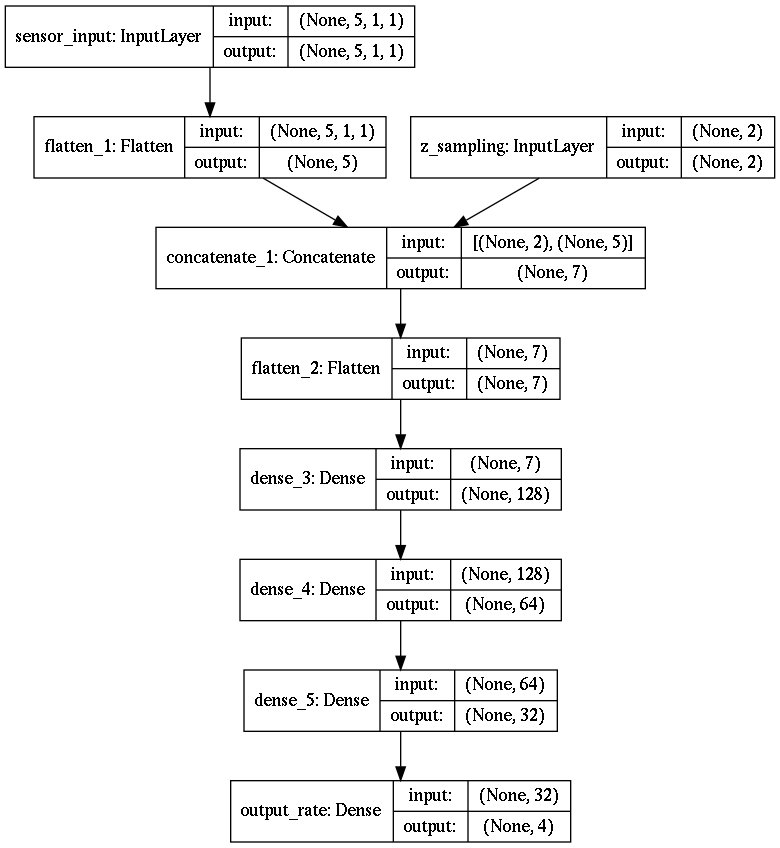}
    \caption{Architecture for the decoder for the classification task}
\end{figure}
\clearpage
\bibliographystyle{unsrt}  

\bibliography{references}

\end{document}